\documentclass{article}
\usepackage{spconf,amsmath,graphicx}
\usepackage{blindtext}
\usepackage[table]{xcolor}
\usepackage{makecell,footmisc,multirow,booktabs,float,tipa}

\title{Leveraging pre-trained representations to improve access to untranscribed speech from endangered languages}

\name{\em{Nay San$^{1,2,6}$\sthanks{These authors contributed equally.} , Martijn Bartelds$^{3}$\footnotemark[1] , Mitchell Browne$^{2,4}$, Lily Clifford$^1$, Fiona Gibson,}\\\em{John Mansfield$^{2,5}$, David Nash$^{2,6}$, Jane Simpson$^{2,6}$, Myfany Turpin$^{2,7}$, Maria Vollmer$^{2,6,8}$,}\\\em{Sasha Wilmoth$^{2,5}$, Dan Jurafsky$^{1}$}\\}

\address{
$^1$Department of Linguistics, Stanford University\\
$^2$ARC Centre of Excellence for the Dynamics of Language\\
$^3$Department of Computational Linguistics, University of Groningen\\
$^4$School of Languages and Cultures, University of Queensland\\
$^5$School of Languages and Linguistics, University of Melbourne\\
$^6$College of Arts and Social Sciences, Australian National University\\
$^7$Sydney Conservatorium of Music, University of Sydney\\
$^8$Department of Linguistics, University of Freiburg\\
{\small\texttt{nay.san@stanford.edu, m.bartelds@rug.nl}}
}

\begin{document}
\maketitle

\begin{abstract}
Pre-trained speech representations like wav2vec 2.0 are a powerful tool for automatic speech recognition (ASR).
Yet many endangered languages lack sufficient data for pre-training such models, or are predominantly oral vernaculars without a standardised writing system, precluding fine-tuning.
Query-by-example spoken term detection (QbE-STD) offers an alternative for iteratively indexing untranscribed speech corpora by locating spoken query terms.
Using data from 7 Australian Aboriginal languages and a regional variety of Dutch, all of which are endangered or vulnerable, we show that QbE-STD can be improved by leveraging representations developed for ASR (wav2vec 2.0: the English monolingual model and XLSR53 multilingual model).
Surprisingly, the English model outperformed the multilingual model on 4 Australian language datasets, raising questions around how to optimally leverage self-supervised speech representations for QbE-STD.
Nevertheless, we find that wav2vec 2.0 representations (either English or XLSR53) offer large improvements (56--86\% relative) over state-of-the-art approaches on our endangered language datasets.
\end{abstract}


\section{Introduction}
\label{sec:intro}

\begin{figure}[ht]
  \centering
  \includegraphics[width=0.76\linewidth]{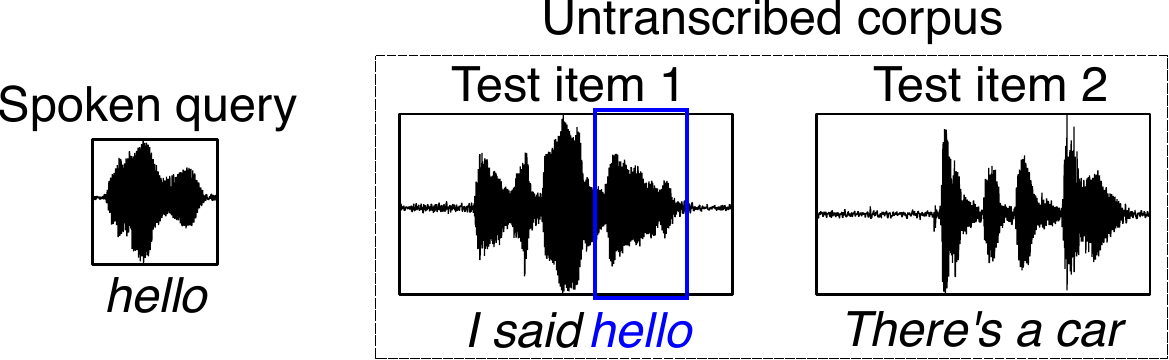}
  \caption{Query-by-example spoken term detection task}
  \label{fig:qbestd}
  \vspace{-1em}
\end{figure}

Recent work has demonstrated the effectiveness of pre-trained speech representations for Automatic Speech Recognition (ASR) \cite{schneiderWav2vecUnsupervisedPretraining2019,baevski2019vq,baevskiWav2vecFrameworkSelfsupervisedtoappear,conneau2020unsupervised}.
Many endangered languages however may lack sufficient amounts of data for pre-training such models.
Further, straightforward applications of ASR systems or standard methods of adaptation (e.g. supervised fine-tuning of pre-trained models) are precluded by the fact that some of these languages predominantly exist as oral vernaculars without a standardised writing system. 

Language documentation efforts to record endangered languages result in a sizable amount of speech data.
While this amount is typically not sufficient to train standard speech recognition systems, it is more than enough to present immediate difficulties for indexing and searching.
These difficulties have a direct impact on how easily such resources may be used for language maintenance and revitalisation activities by many interested parties, from speakers within the communities, to language teachers, to linguists.

In this paper, we focus on query-by-example spoken term detection (QbE-STD), which, as illustrated above in Figure~\ref{fig:qbestd}, is a long-standing speech information retrieval task of finding all regions within a corpus of audio documents where a spoken query term occurs \cite{myersInvestigationUseDynamic1980,rohlicek1995Word,fiscusResults2006Spoken2007,rodriguez-fuentesHighperformanceQuerybyexampleSpoken2014}.
For language documentation projects, QbE-STD can be leveraged to iteratively and interactively index a corpus of untranscribed speech using a recorded list of common words as an initial set of queries \cite{birdSparseTranscriptiontoappear,leferrandEnablingInteractiveTranscription2020}.

Using data from 7 Australian Aboriginal languages and Gronings, a variant of Low Saxon (one of the minority languages spoken in the Netherlands), we show that QbE-STD performance on these languages can be appreciably improved by using a large self-supervised neural model trained only on English speech (wav2vec 2.0, trained on the 960-hour LibriSpeech dataset: LS960).
In the worst of cases, retrieval performance improved 56--86\% from only 27--28\% of queries being retrievable to 42--52\% when using representations from the wav2vec 2.0 LS960 model.
This is a tolerable operating range, given the alternative is browsing untranscribed audio in near real-time.

Our main contributions are as follows: 1) a demonstration of how a pre-trained model can be used for QbE-STD where there is insufficient data for training feature extractors in the unrelated target language(s); 2) an evaluation of the method using real data from ongoing language documentation projects; and 3) an error analysis and visualisations of phonetic features encoded in the pre-trained representations to guide how this approach could be further refined.

\begin{table*}[!t]

  \caption{Characteristics of evaluation datasets. Parentheticals indicate speaker compositions, and total length of test audio in minutes. Same, overlap, or different indicate whether the queries and test items shared speakers and/or recording conditions.}
  \label{tab:datasets}
  \centering
  \begin{tabular}{llcccc}
    \toprule
    \multirow{2}{*}{Dataset} & \multirow{2}{*}{Language} &  \multirow{2}{*}{\makecell{Number of queries\\(Speakers)}} & \multirow{2}{*}{\makecell{Number of test items\\(Speakers; Duration)}} & \multicolumn{2}{c}{Query vs. Test items} \\ \cmidrule(l){5-5}\cmidrule(l){6-6}
     & & & & Speakers & Recordings \\
    \midrule
    gbb-pd   & Kaytetye         & 397 (1F)     & 397 (1F; 16 mins)   & Same  &  Same  \\
    wrm-pd   & Warumungu        & 383 (1F)     & 383 (1F; 20 mins)   & Same & Same  \\

    wrl-mb   & Warlmanpa        & 23  (1F)     & 162 (1F; 11 mins)   & Same & \cellcolor{gray!60}Different  \\
    gup-wat  & Kunwinjku        & 50  (5M)      & 725 (5M, 2F; 32 mins)   & \cellcolor{gray!20}Overlap &  \cellcolor{gray!20}Overlap  \\

    gbb-lg   & Kaytetye         & 189 (1F)     & 809 (8F, 4M; 45 mins)   & \cellcolor{gray!20}Overlap & \cellcolor{gray!60}Different  \\

    pjt-sw01 & Pitjantjatjara   & 30  (1F)     & 320 (5F; 38 mins)   & \cellcolor{gray!60}Different & \cellcolor{gray!60}Different  \\
    mwf-jm   & Murrinhpatha     & 37  (1F, 1M) & 259 (3F, 2M; 13 mins)   & \cellcolor{gray!60}Different & \cellcolor{gray!60}Different  \\
    wbp-jk   & Warlpiri         & 24  (1F)     & 198 (12F; 6 mins)    & \cellcolor{gray!60}Different & \cellcolor{gray!60}Different  \\

    \cmidrule(lr){1-6}
    gos-kdl  & Gronings         & 83 (2F, 3M)           & 430 (2F, 2M; 23 mins)   & \cellcolor{gray!20}Overlap & \cellcolor{gray!20}Overlap  \\
    eng-mav  & English          & 100 (1M)     & 526 (1M; 77 mins)   & \cellcolor{gray!60}Different & \cellcolor{gray!60}Different  \\
    \bottomrule
  \end{tabular}
  
\end{table*}

\section{Background}

\subsection{Query-by-example spoken term detection}

A conventional QbE-STD system consists of two broad stages.
In the first stage, features are extracted from the query and test items.
The second stage involves calculating a detection score of how likely a query occurs in a test item for each pair of query and test item, typically using a Dynamic Time Warping (DTW) based template matching.
Like many conventional speech processing tasks, a noise-robust and speaker-invariant feature extraction method is a key component of a performant QbE-STD system.
Indeed, the experiments in \cite{leferrandEnablingInteractiveTranscription2020} showed that while QbE-STD was promising for facilitating searches on speech from language documentation efforts, considerably lower retrieval performance was observed when the speaker of the query and that of the test item were different.

Recent advances in QbE-STD have involved using neural networks for either the feature extraction stage, e.g.~using bottleneck features (BNF) \cite{menonFeatureExplorationAlmost2018},\footnote{Bottleneck features are neural features extracted from a hidden layer that has a lower dimensionality compared to the surrounding layers.} or the detection classification stage, e.g.~using a convolutional neural network (CNN) \cite{ramCNNBasedQuery2018}, or forgoing conventional stages entirely by using an end-to-end system \cite{ramNeuralNetworkBased2020}.
However, not all languages benefit equally from these advances.
While the neural systems (BNF+CNN and end-to-end) in \cite{ramNeuralNetworkBased2020} outperformed the BNF+DTW system for larger European languages from the SWS2013 QbE-STD benchmark dataset, the highest performing system for all of the smaller African languages was the BNF+DTW system. 
In other words, for languages for which little or no labelled QbE-STD datasets exists to train neural classifiers, the DTW-based approach remains state-of-the-art.
Thus in this paper we investigate whether QbE-STD systems using self-supervised speech representations in place of bottleneck features result in better retrieval performance for low- and zero-resource target languages.

\subsection{Pre-trained self-supervised speech representations}

Training a bottleneck feature extractor typically requires speech labelled at the phone level, which is costly to generate even for major languages.
There has therefore been interest in deriving useful speech representations from audio alone \cite{chen2018almost,kahn2020self}.
One such framework is wav2vec (w2v) \cite{schneiderWav2vecUnsupervisedPretraining2019}, a CNN that takes raw audio as input and learns representations useful for differentiating true future audio samples from randomly sampled false ones. 

The wav2vec 2.0 (w2v2) framework involves appending a quantiser module (to discretise the CNN output) and a 24-layer Transformer network, which uses the discretised outputs to learn better contextualised representations \cite{baevskiWav2vecFrameworkSelfsupervisedtoappear}.
Typically, representations extracted from the final layers of a Transformer network tend to be more suited to the original training task than its middle layers, which are better suited for downstream tasks \cite{tenney-etal-2019-bert}.

Experiments in \cite{bartelds2020neural} investigated which of the 24 w2v2 Transformer layers may be best suited for automatic pronunciation scoring of non-native English speech.
Similar to QbE-STD, the task of pronunciation scoring is a 2-stage process: features are extracted from native and non-native speech samples of the same read text, and then a DTW-based distance is calculated (lower distance indicates closer pronunciations).
Results from \cite{bartelds2020neural} showed that distances between native and non-native speech samples had a strong negative correlation with native speaker ratings of the same samples (\emph{r} = -0.85; lower distances correspond to higher ratings) when computed using representations extracted from the 10th Transformer layer.
In other words, the representations of the middle layers of the w2v2 Transformer network appear to be well-suited for comparing phonetic characteristics of speech samples while being robust to recording conditions and speaker characteristics.

\section{Experimental setup}

As mentioned in the introduction, previous experiments in \cite{leferrandEnablingInteractiveTranscription2020} trialling QbE-STD with language documentation data showed that occurrences of a query were less likely to be detected when the query and test item being compared were spoken by different speakers.
Our own pilot experiments with Mel-frequency cepstral coefficients (MFCC) and bottleneck features (BNF), which we report here as baselines, had also shown that even when the speaker was the same, acoustic differences arising from recording equipment and/or environment differences can result in similar drops in performance.
Motivated by these issues, we examined whether performance improved with features extracted from various stages of the w2v2 English and multilingual models for QbE-STD using datasets in which the speakers and/or recording conditions between the query and test items were the same, overlapping, or different.
We examined both the English and multilingual w2v2 models as earlier work had shown that for keyword spotting (i.e.~QbE-STD with a closed query set) in West African languages the English w2v model performed on par with a w2v model trained on 10 West African languages \cite{doumbouya2021usingradio}.

\subsection{Datasets}

We curated data from English, Gronings, and 7 Australian Aboriginal languages.
Gronings and all of the Australian languages are currently classified as vulnerable or endangered \cite{eberhard_ethnologue_2020}. 
For these languages, the audio were sourced from various language documentation projects (e.g., audio for dictionaries, audio book of short stories) and thus constitute in-domain test data for the QbE-STD systems evaluated.

We gathered for each language queries from 1-5 speakers consisting of words and multi-word expressions and test items consisting of sentences and longer phrases from 1-12 speakers.
Each dataset varied according to whether the query and test items shared speaker(s) and/or recording conditions, as indicated in Table \ref{tab:datasets}.
For example, for gbb-lg, the 189 queries were sourced from audio produced in a recording studio by a single female speaker AR.
The 809 test items were sourced from various field recordings from 12 speakers (8F, 4M: one of whom was AR).
Thus, for gbb-lg, the queries and test items overlapped in speakers but differed in recording conditions.

\subsection{Pre-trained models}

In our feature extraction procedure, we use two w2v2-based models, namely the English monolingual model described in \cite{baevskiWav2vecFrameworkSelfsupervisedtoappear} and the multilingual `XLSR53' model described in \cite{conneau2020unsupervised}.
The English monolingual model is pre-trained on the unlabelled Librispeech dataset (LS960), which contains 960 hours of clean and noisy English speech from audio books.
The multilingual model is pre-trained on 56,000 hours of data from 53 different languages (amount of data varies per language), and includes African, Asian, and European languages obtained from audio books, conversational telephone speech, and read speech.
The architecture of both models is the same, except for the quantisation of the encoder representations.
XLSR53 learns a single set of quantised speech representations on the basis of the encoder output.
This set was subsequently shared across languages, which was shown to be effective for learning self-supervised multilingual speech representations.

\subsection{Procedure}

To create ground truth labels, we first paired each query's text transcription with that of each test item within a given dataset.
Then for each pair, we used regular expressions to determine whether the query occurred in the test item.
All audio data from various sources were standardised to mono 16-bit PCM at 16 kHz (sample rate required by w2v2).
For MFCC and bottleneck features (BNF), the audio was further downsampled to 8 kHz at extraction time to meet the requirements of the respective feature extractors.

For baseline comparisons, we extracted MFCC and BNF representations using the Shennong library,\footnote{https://docs.cognitive-ml.fr/shennong/} which provides a Python interface to common feature extraction routines.
For MFCC, we used the Kaldi feature extractor which returns for each time frame 13 MFCC features with their first and second derivatives (39 features total). 
For BNF, we used the BUT/Phonexia feature extractor which returns for each time frame 80 activation values from a bottleneck layer originally trained for phone classification on the 17 languages of the IARPA Babel dataset \cite{silnovaPhonexiaBottleneckFeature2018}.
For w2v2 features (both from the English monolingual model and XLSR53 multilingual model), we adapted the feature extraction code from \cite{bartelds2020neural},\footnote{https://github.com/Bartelds/neural-acoustic-distance} which extracts outputs from the Encoder CNN (E), the Quantiser module (Q), or any one of the 24 layers of the Transformer network (T01--T24).
Thus, in total we extracted features using 54 different methods for each of the 10 datasets.

As our primary interest is in comparing different feature extraction methods, we implemented a DTW-based detection stage based on a state-of-the-art QbE-STD system \cite{rodriguez-fuentesHighperformanceQuerybyexampleSpoken2014}.
Given two feature matrices, of a query $Q$ and of a test item $T$, we calculated standardised Euclidean distances between each time frame in $Q$ and $T$.
To make distances between different pairs of queries and test items comparable, these distances were range normalised to [0, 1].
A window the size of the query was moved along the normalised distance matrix and a DTW-based distance between the query and subpart of the test item was calculated at each step.
The score for how likely a query occurs in a test item was then calculated as 1 minus the minimum distance.

We used the Maximum Term Weighted Value (MTWV) to evaluate each of our QbE-STD systems (e.g.~MFCC+DTW, w2v2-E+DTW, etc.).
The MTWV ranges from 0, indicating a system that simply returns nothing, to 1, indicating a perfect system detecting all relevant instances with no false positives.
Using the NIST STDEval tool,\footnote{https://www.nist.gov/itl/iad/mig/tools} we calculated MTWVs with the suggested NIST costs (false positive: 1, false negative: 10) and an empirical prior of 0.0278 (true positive rate averaged across all 10 datasets).
With these costs, a MTWV of 0.48 indicates a system that correctly detects 48\% of all queries searched, while producing at most 10 false positives for each true positive correctly retrieved \cite{whiteUsingZeroresourceSpoken2015}.

For significance testing, we computed the MTWV for each query in a given dataset and performed a one-sided paired t-test to examine whether the MTWVs obtained for the queries using one feature extraction method were significantly greater than those obtained using another method.

With the exception of the raw audio and transcriptions from the Australian language documentation projects for which we do not have permissions to release openly, we have made all pilot experiments results, experiment code and analyses available on a GitHub repository, and placed the Gronings data and all experimental artefacts in Zenodo archives.\footnote{All linked on: https://github.com/fauxneticien/qbe-std\_feats\_eval\label{ghlink}}
The English data can be obtained from the Mavir corpus \cite{sandoval2012mavir}.\footnote{http://www.lllf.uam.es/ING/CorpusMavir.html}

\begin{figure*}[ht!]
  \centering
  \includegraphics[width=0.97\linewidth]{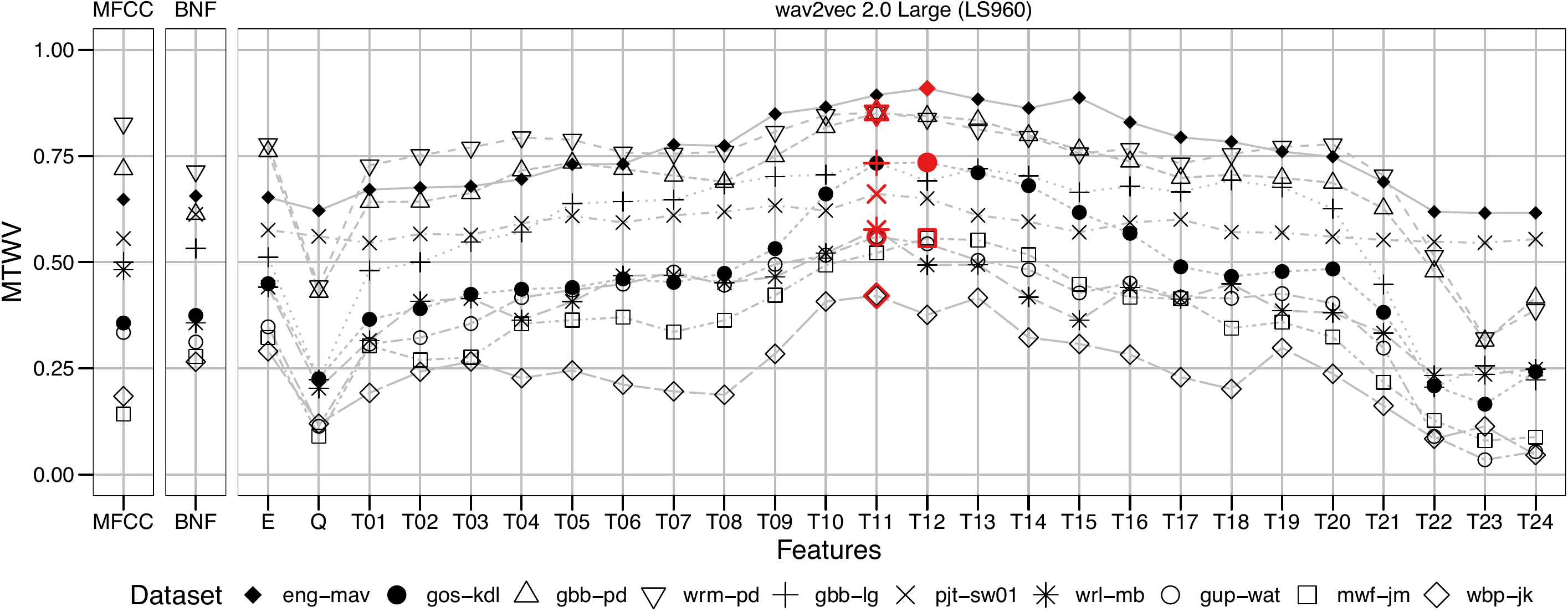}
  \includegraphics[width=0.97\linewidth]{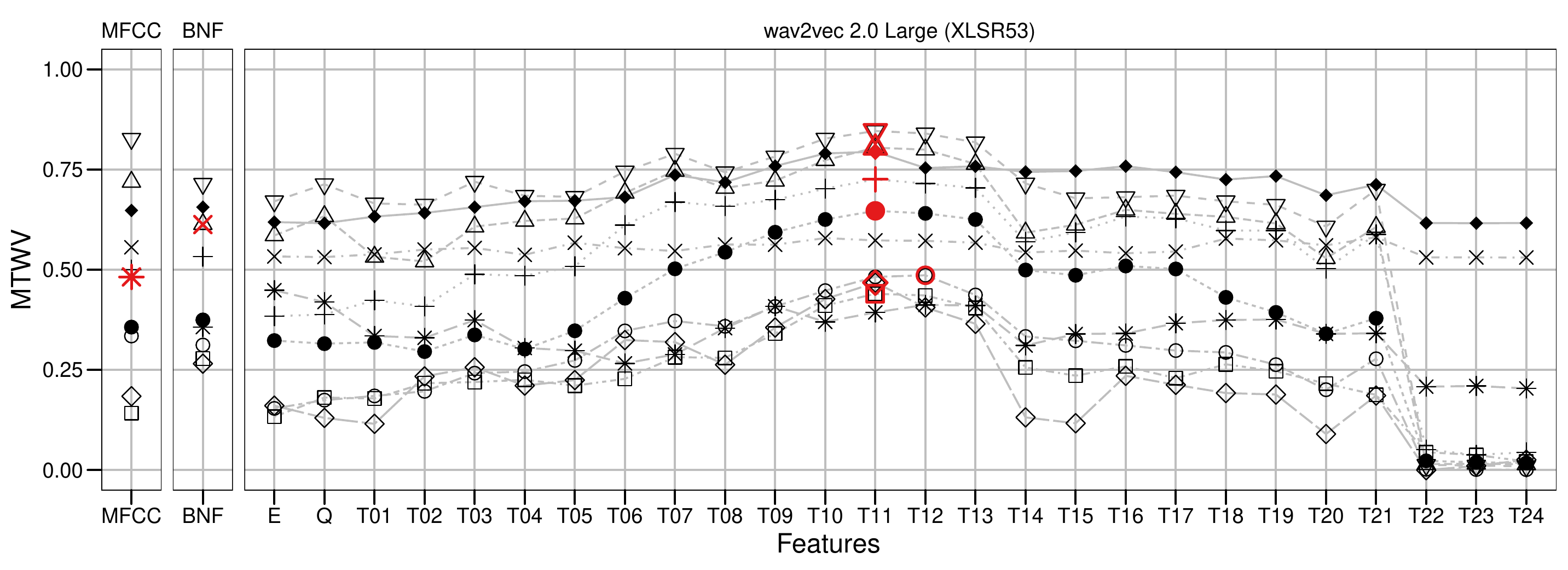}
  \caption{Maximum Term Weighted Values achieved on a QbE-STD task using various speech representations: baselines (MFCC; bottleneck features: BNF) and from layers (E; Q; T01-24) of two wav2vec 2.0 models (English monolingual: LS960; multilingual: XLSR53). Shapes indicate one of 10 datasets; red-coloured/largest shape indicates highest score achieved on dataset.}
  \label{fig:mtwv}
\end{figure*}

\begin{table*}[!ht]

  \caption{Results of one-sided \emph{t}-tests for the alternative hypothesis ($H_1$) that the Maximum Term Weighted Value (MTWV) obtained for each query in a given dataset is greater using different speech representations: baseline bottleneck features (BNF), and Transformer layer 11 (T11) of two wav2vec 2.0 models (English monolingual: LS960; multilingual: XLSR53).}
  \label{tab:ttests}
  \centering
  \begin{tabular}{lccccccc}
    \toprule
    \multirow{2}{*}{\makecell{Dataset\\($N$ queries)}} & \multicolumn{3}{c}{MTWV: mean (standard deviation)} & \multicolumn{3}{c}{One-sided paired \emph{t}-test (\emph{t}-value, \emph{p}-value)} \\ \cmidrule(l){2-4}\cmidrule(l){5-7}
     & a. BNF & b. LS960-T11 & c. XLSR53-T11 & $H_1$: b. $>$ a. & $H_1$: c. $>$ a. & $H_1$: b. $>$ c. \\
    \midrule
    gbb-pd (397)  & 0.612 (0.383)        & 0.849 (0.248)     & 0.803 (0.309)  & (14.1, .000)  &  (11.1, .000) & (3.8, .000) \\
    wrm-pd (383)  & 0.713 (0.361)        & 0.853 (0.256)     & 0.843 (0.260)  & (8.7, .000) & (8.2, .000) & \cellcolor{gray!60}(0.7, .236) \\
    wrl-mb (23)   & 0.352 (0.213)        & 0.573 (0.309)     & 0.379 (0.308)  & (3.0, .003) & \cellcolor{gray!60}(0.3, .372) & (3.4, .001) \\
    gup-wat (50)  & 0.313 (0.271)        & 0.561 (0.266)     & 0.483 (0.292)  & (6.6, .000) &  (3.9, .000) & (2.7, .005) \\
    gbb-lg (189)  & 0.533 (0.360)        & 0.733 (0.338)     & 0.719 (0.327)  & (6.7, .000) & (6.6, .000) & \cellcolor{gray!60}(0.5, .299) \\
    pjt-sw01 (30) & 0.607 (0.163)        & 0.660 (0.210)     & 0.572 (0.207)  & \cellcolor{gray!60}(1.4, .080) & \cellcolor{gray!60}(-0.9, .082) & (2.2, .020)  \\
    mwf-jm (37)   & 0.279 (0.330)        & 0.515 (0.365)     & 0.435 (0.346)  & (4.8, .000)  & (2.4, .010) & \cellcolor{gray!60}(1.3, .102) \\
    wbp-jk (24)   & 0.267 (0.353)        & 0.422 (0.356)     & 0.468 (0.325)  & (1.9, .037)  & (2.6, .009) & \cellcolor{gray!60}(-0.6, .732)  \\
    \cmidrule(lr){1-7}
    gos-kdl (83)  & 0.372 (0.335)        & 0.728 (0.281)     & 0.647 (0.306)  & (9.8, .000)  & (8.2, .000) & (2.9, .002) \\
    eng-mav (100) & 0.656 (0.078)        & 0.894 (0.222)     & 0.794 (0.248)  & (11.1, .000) & (5.7, .000) & (3.9, .000)  \\
    \bottomrule
  \end{tabular}
  
\end{table*}

\begin{figure*}[!ht]
  \centering
  \includegraphics[width=0.75\linewidth]{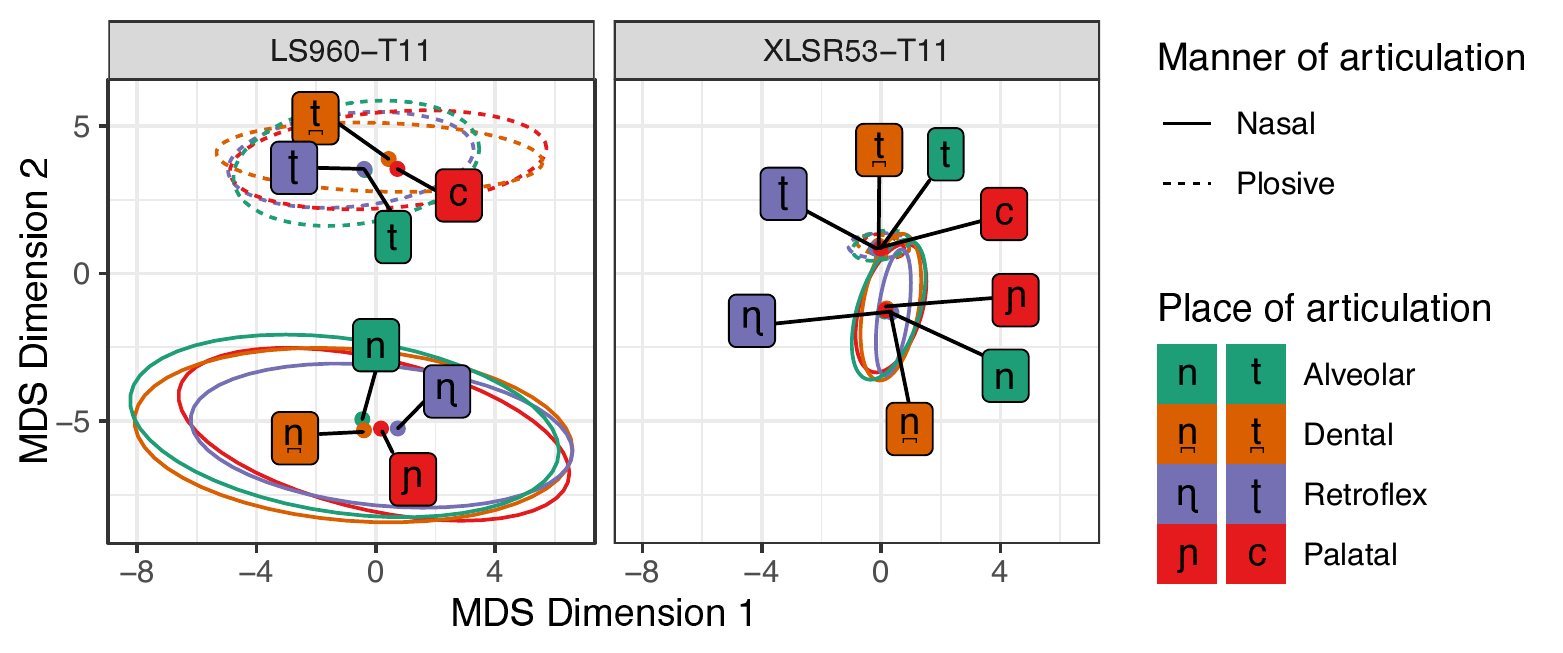}
  \caption{Multidimensional scaling (MDS) visualisations of features extracted from Kaytetye consonants (data from \cite{harvey2015contrastive}) using Transformer layer 11 (T11) of wav2vec 2.0 (English monolingual: LS960; multilingual: XLSR53), averaged over the duration of the consonant. Ellipses represent 95\% confidence intervals and text labels represent means.}
  \label{fig:ellipses}
\end{figure*}

\section{Results and Discussion}
\label{sec:results}

Figure \ref{fig:mtwv} displays the MTWVs achieved by QbE-STD systems using various feature extraction methods on each of the 10 datasets.
As expected, dataset characteristics had a clear effect on QbE-STD performance.
For our baselines, retrieval performance using MFCC was highest on the datasets having the same speaker and recordings between the queries and test items (wrm-pd: 0.83; gbb-pd: 0.71), lower on those with overlapping speakers (e.g.~wrl-mb: 0.48) or recordings (e.g.~gup-wat: 0.33), and lowest when both were entirely different (e.g.~wbp-jk: 0.18, mwf-jm: 0.14).
The same effects were also seen using BNF, which improved on different-speaker/recordings datasets (e.g.~wbp-jk: 0.27), though at the cost of worsening on those with same-speaker/recordings (e.g.~wrm-pd: 0.71).

Using outputs from a middle Transformer layer of a w2v2 model resulted in increased performance on nearly all our datasets.
For 9 of 10 datasets, the MTWVs obtained using outputs of the Transformer layer 11 from the English monolingual model were significantly greater than those obtained using BNF, as indicated by the $t$-tests in Table \ref{tab:ttests} (b. LS960-T11 $>$ a. BNF).
Similarly, the MTWVs obtained using Transformer layer 11 of the multilingual model were significantly greater than those using BNF for 8 of 10 datasets (Table \ref{tab:ttests}: c. XLSR53-T11 $>$ a. BNF).
Overall, the best performing representations were those from the Transformer layer 11 of the English w2v2 model, which also significantly outperformed those from the equivalent layer of the multilingual w2v2 model for 4 of the 8 Australian language datasets (Table \ref{tab:ttests}: b. LS960-T11 $>$ c. XLSR53-T11).
On the two most challenging Australian language datasets with different speakers and recordings, performance increased 56-86\% using T11 features of the English model (wbp-jk: 0.42, mwf-jm: 0.52) over the best achievable baseline scores using BNF (wbp-jk: 0.27, mwf-jm: 0.28); n.b. the English and multilingual models did not differ significantly on these two datasets.
These results demonstrate the effectiveness of contextualised representations learned by the Transformer network for comparing the phonetic characteristics of speech samples while being robust to recording conditions and speaker characteristics.

To better understand the representations of the Transformer layer 11 of the English w2v2 model, we undertook an error analysis of the erroneous retrievals by the QbE-STD system on the Australian languages when using these features.
We focused on `unretrievable` queries, defined as those for which no true matches were present in the top 5 results returned for that query, and examined the differences in phonetic transcriptions between the query and its top match.
This analysis revealed that for the majority of these queries (89/119 = 75\%), there were 4 or fewer segmental differences between the transcriptions of the query and its top match, and that, notably, the query and its top match typically shared a manner template.
For example, for the Kaytetye query [\textipa{a\:nanp@}] ‘medicinal sap’, its top match was the word [\textipa{an@nk@}] ‘to sit’, both of which share the VNVNTV template (where V represents a vowel, N a nasal, and T a plosive).
In other words, the representations of the w2v2 English model do not appear to be sufficiently fine-grained to differentiate between segments differing primarily in place of articulation, especially for those that do not occur contrastively in English (e.g. retroflex nasal [\textipa{\:n}] vs. alveolar nasal [n]).

\subsection{Exploring wav2vec 2.0 features}

Based on the findings of our error analysis, we examined whether differences in the way phonetic information is encoded by the w2v2 English and the multilingual models could explain differences in their QbE-STD performance.
We used the Transformer layer 11 of the English and multilingual models to extract features from Kaytetye consonants (from audio stimuli collected by \cite{harvey2015contrastive}).
Figure \ref{fig:ellipses} displays the dissimilarities between each of the Kaytetye consonant types in the 1024-dimensional space of the two w2v2 models (averaged across the duration of the consonant and reduced to 2 dimensions through multidimensional scaling).
For considerations of space and visual clarity, we focus on the nasals and plosives.

Despite being given no phone labels at training time, the w2v2 models learn to encode broad manner of articulation classes within its representations, as evidenced by ellipses forming two clusters for nasals (solid lines) and plosives (dashed lines) in Figure \ref{fig:ellipses}.
In line with findings from our qualitative error analysis, features from the w2v2 English model under-differentiate place of articulation contrasts, as evidenced by the overlapping ellipses within each of the nasal and plosive manners of articulation in Figure \ref{fig:ellipses}. 
Compared to the English model, the representations of the Kaytetye consonants by the multilingual model overlap to a larger degree.
This result suggests that representations extracted from a model trained on a single language or a set of phonologically similar languages may be more beneficial for QbE-STD than a large multilingual model such as XLSR53 trained on a diverse set of 53 languages --- especially when supervised fine-tuning in the target languages is not a readily available option.

\section{Conclusions}

We compared the performance of 54 feature extraction methods for QbE-STD using data from 9 languages, eight of which are vulnerable or endangered.
Our results showed that the middle layers of the wav2vec 2.0 Transformer network provide a noise-robust and speaker-invariant representation of speech sounds that can be leveraged for QbE-STD with low resource languages, performing significantly better than state-of-the-art approaches using bottleneck features.
An error analysis and a preliminary investigation of the information encoded in the representations learned by the wav2vec 2.0 Transformer network suggested that the representations extracted from the English model may provide better features for QbE-STD in unrelated languages than the multilingual model in cases where important phonological distinctions in those languages are not as distinctively encoded by the multilingual model.

\section{Acknowledgements}

We thank all those who contributed data which made our experiments possible: Steven Bird (Kunwinjku); Samantha Disbray (Warumungu); Jennifer Green and Angela Harrison (Warlpiri); Alison Nangala Ross (Kaytetye); the Klunderloa project/CGTC, Riemke Bakker, Eltje Doddema, Geesjen Doddema, Henk Scholte and Olaf Vos (Gronings).

\newpage

\bibliographystyle{IEEEbib}
\bibliography{sources}

\end{document}